\documentclass[pdflatex,sn-mathphys-num]{sn-jnl}

\usepackage{graphicx}%
\usepackage{multirow}%
\usepackage{amsmath,amssymb,amsfonts}%
\usepackage{amsthm}%
\usepackage{mathrsfs}%
\usepackage[title]{appendix}%
\usepackage{xcolor}%
\usepackage{textcomp}%
\usepackage{manyfoot}%
\usepackage{booktabs}%
\usepackage{algorithm}%
\usepackage{algorithmicx}%
\usepackage{algpseudocode}%
\usepackage{listings}%

\theoremstyle{thmstyleone}%
\theoremstyle{thmstyletwo}%

\theoremstyle{thmstylethree}%

\begin{document}

\title[Article Title]{DF-SLAM: Dictionary Factors Representation for High-Fidelity Neural Implicit Dense Visual SLAM System}


\author[1]{\fnm{Weifeng} \sur{Wei}}\email{416100220248@email.ncu.edu.cn}

\author[2]{\fnm{Jie} \sur{Wang}}\email{8008121372@email.ncu.edu.cn}

\author[1]{\fnm{Shuqi} \sur{Deng}}\email{6108122062@email.ncu.edu.cn}

\author*[2]{\fnm{Jie} \sur{Liu}}\email{ndliujie@ncu.edu.cn}

\affil[1]{\orgdiv{School of Information Engineering}, \orgname{Nanchang University}, \orgaddress{\city{Nanchang}, \postcode{330031}, \country{China}}}

\affil[2]{\orgdiv{School of Software}, \orgname{Nanchang University}, \orgaddress{\city{Nanchang}, \postcode{330103}, \country{China}}}


\abstract{We introduce a high-fidelity neural implicit dense visual Simultaneous Localization and Mapping (SLAM) system, termed DF-SLAM. In our work, we employ dictionary factors for scene representation, encoding the geometry and appearance information of the scene as a combination of basis and coefficient factors. Compared to neural implicit dense visual SLAM methods that directly encode scene information as features, our method exhibits superior scene detail reconstruction capabilities and more efficient memory usage, while our model size is insensitive to the size of the scene map, making our method more suitable for large-scale scenes. Additionally, we employ feature integration rendering to accelerate color rendering speed while ensuring color rendering quality, further enhancing the real-time performance of our neural SLAM method. Extensive experiments on synthetic and real-world datasets demonstrate that our method is competitive with existing state-of-the-art neural implicit SLAM methods in terms of real-time performance, localization accuracy, and scene reconstruction quality. Our source code is available at \href{https://github.com/funcdecl/DF-SLAM} {https://github.com/funcdecl/DF-SLAM}.}

\keywords{Neural implicit dense visual SLAM; Dictionary factors; Feature integration rendering}



\maketitle

\section{Introduction}\label{sec1}

The primary goal of dense visual Simultaneous Localization and Mapping (SLAM) is to estimate the pose of visual sensor (such as RGB-D camera) while constructing dense 3D maps of unknown environments. It has wide applications in the fields of autonomous driving, robotics, and virtual/augmented reality. Traditional visual SLAM methods such as \cite{ORB-SLAM, ORB-SLAM2, ORB-SLAM3} employ multi-view geometry to achieve robust camera tracking and represent scene maps using point clouds. However, this discrete form of scene representation typically requires substantial memory resources and exhibits heightened sensitivity to environmental conditions.

The emergence of Neural Radiance Fields (NeRF) \cite{NeRF} demonstrates that neural networks have powerful spatial continuous representation capabilities, giving full play to their advantages in dense visual SLAM. iMAP \cite{iMAP} utilizes a single Multi-Layer Perceptron (MLP) as scene representation, constructing a loss function via volume rendering to iteratively optimizes the scene representation and estimates camera poses. NICE-SLAM \cite{NICE-SLAM} introduces hierarchical feature grids and pre-trained tiny-MLPs as scene representation. This can locally update the constructed map, avoiding catastrophic forgetting. Afterward, ESLAM \cite{ESLAM} and Co-SLAM \cite{Co-SLAM} utilize axis-aligned feature planes and hash grid as scene representation, respectively. Compared to NICE-SLAM \cite{NICE-SLAM}, they have a lower memory footprint growth rate.

However, the above methods have some issues. For scene representation, the model size of NICE-SLAM \cite{NICE-SLAM} grows cubically with the range of scene map, and higher resolution grids are required to accurately represent the fine details in the scene, which further increases memory usage and training costs. Despite ESLAM \cite{ESLAM} only has a quadratic memory footprint growth rate, it remains sensitive to the size of the map. In addition, Axis-aligned feature planes decompose scene information along coordinate axes. If critical features or structures are not aligned with the main coordinate axes, axis-aligned transformations struggle to capture them effectively. Due to the hash grid is prone to hash collisions at fine scales, Co-SLAM \cite{Co-SLAM} faces challenges in reconstructing detailed regions of the scene. Furthermore, Co-SLAM \cite{Co-SLAM} requires a customized CUDA framework, which exhibits limited scalability when addressing downstream tasks in neural SLAM. For rendering, these methods use ray marching through query and volume rendering space points for scene color rendering. This process requires querying the RGB values of all points along the ray, resulting in huge MLP query costs and affecting the real-time performance of the neural SLAM method.

To further balance accuracy and performance, we are inspired by \cite{tensorf, factor-field} to utilize dictionary factors for neural implicit dense visual SLAM scene representation, and propose employing a neural feature rendering method similar to \cite{fr}, namely feature integration rendering to improve the real-time performance of neural implicit SLAM. This enables us to reconstruct detailed scene maps in real-time and unaffected by the memory growth associated with incremental reconstruction. In general, our contributions include the following:
\begin{itemize}
\item We propose a neural implicit dense visual SLAM model based on dictionary factors representation, demonstrating higher-fidelity detail reconstruction and a more compact model memory footprint in large-scale scenes.
\item To address insufficient real-time performance due to high MLP query costs in volume rendering in previous neural SLAM methods, we introduce an efficient rendering approach using feature integration. This improve performance without relying on customized CUDA framework.
\item We conduct extensive experiments on both synthetic and real-world datasets to validate our design choices, showing competitive speed and quality compared to baselines.
\end{itemize}

\section{RELATED WORKS}\label{sec2}
\subsection{Traditional and Learning-based Dense Visual SLAM} \label{sec:2.1}
Most of the existing dense visual SLAM methods follow the tracking and mapping architecture proposed by PTAM \cite{PTAM}. DTAM \cite{DTAM} introduces the first dense visual SLAM system, utilizes cost volume for scene representation and employs a dense per-pixel approach for tracking and mapping. KinectFusion \cite{kinectfusion} proposes a pioneering method for real-time scene reconstruction using RGB-D camera as sensor. They employ Iterative Closest Point (ICP) for tracking and utilize TSDF-Fusion to update scene geometry for mapping. BAD-SLAM \cite{BAD-SLAM} utilizes dense surfels for representing scene. These surface elements can be efficiently updated through direct Bundle Adjustment (BA). Some recent works \cite{code-slam, droid-slam, node-slam, deep-slam} further improve the accuracy and robustness of dense visual SLAM system compared to traditional methods by combining deep neural networks with traditional visual SLAM framework.

\subsection{Neural Implicit Representations} \label{sec:2.2}
NeRF \cite{NeRF} has garnered extensive research interest in novel view synthesis and 3D reconstruction due to its continuous scene representation and memory efficiency. However, this MLP-based neural implicit representation requires lengthy training time. Subsequent works have adopted hybrid scene representations, encoding scene information into features and anchoring these features onto specific data structures such as octrees \cite{neural, plenoctrees}, voxel grids \cite{DVGO, vox-surf}, tri-planes \cite{EG3D}, and hash grids \cite{iNGP}. This approach significantly accelerates training speeds at the cost of increased memory usage. These methods provide support for the real-time performance of neural implicit dense visual SLAM methods.

\subsection{Neural Implicit Dense Visual SLAM} \label{sec:2.3}
iMAP \cite{iMAP} first proposes a neural implicit dense visual SLAM system that uses a single MLP for tracking and mapping. While it exhibits efficient memory usage, it faces catastrophic forgetting in large-scale scenes. NICE-SLAM \cite{NICE-SLAM} introduces hierarchical feature grids for scene representation to avoid forgetting and achieve large-scale scenes reconstruction. ESLAM \cite{ESLAM} utilizes axis-aligned feature planes to reduce memory footprint growth rate. Vox-Fusion \cite{Vox-Fusion} employs a sparse octree for scene representation and does not require a predefined scene bounding box. Co-SLAM \cite{Co-SLAM} employs hash grid for scene representation and one-blob encoding for hole-filling. Point-SLAM \cite{point-slam} adopts neural point clouds for better 3D reconstruction. More recently, GS-SLAM \cite{GS-SLAM} combines 3D Gaussian Splatting representation \cite{3dgs} with the neural implicit SLAM framework, demonstrating faster rendering speed and photorealistic scene reconstruction.

\section{METHOD}\label{sec3}
\begin{figure*}
	\centering
	\includegraphics[width=13cm]{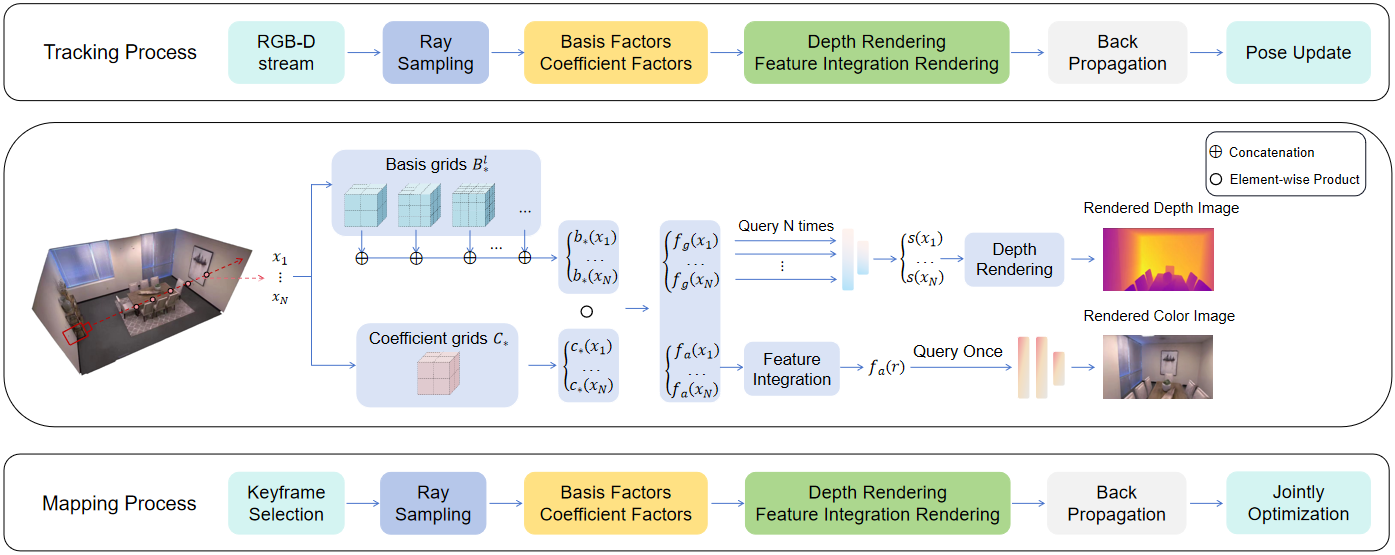}
	\caption{\textbf{Overview.} 1) Scene Representation: We use two different sets of factor grids to represent the scene geometry and appearance respectively. To simplify our overview, we use symbol ${*}$ to denote both geometry $g$ and appearance $a$, e.g., $b_{*}(x_i)$ can be either $b_g(x_i)$ or $b_a(x_i)$. For sample points along the ray, we query the basis and coefficient factors for depth and feature integration rendering. 2) Mapping process: Jointly optimize scene representation and camera poses 3) Tracking process: By minimizing the losses, each input camera pose is updated.} 
   \label{fig:1}
\end{figure*}

Fig.\ref{fig:1} shows an overview of our work. Section \ref{sec:3.1} describes our scene representation in detail. Section \ref{sec:3.2} introduces feature integration color rendering, and Section \ref{sec:3.3} introduces the details of tracking, mapping and the loss functions.

\subsection{Dictionary Factors Representation}\label{sec:3.1}
Different from \cite{NICE-SLAM,ESLAM, Co-SLAM} which directly encode scene information into features, we encode scene information into basis and coefficient factors, utilizing the combination of these factors as the features. Since the geometry of a scene generally converges faster than its appearance, utilizing a unified factors model to jointly represent both can readily lead to the forgetting of geometry learning in the scene. This reduces pose estimation robustness. Therefore, we adopt different basis and coefficient factors to represent the geometry and appearance of the scene respectively. The basis and coefficient factors are implemented via learnable tensor grids. For the basis factors, we implement them using multi-level tensor grids $B^l_g$ and $B^l_a$ , where each level of resolution increases linearly, this enables us to model the different frequencies of information in the scene more accurately. While for the coefficient factors, we use single level tensor grids $C_g$ and $C_a$. Specifically, For $N$ sampled points $x_{i}=\mathbf{o}+z_{i} \mathbf{d},\quad i \in\{1, \ldots, N\}$ along the ray, where $\mathbf{o}$ denotes camera origin and $z_{i}$ corresponds to depth value of sampled point $x_{i}$. we query the geometry basis factor $b_g(x_i)$ and geometry coefficient factor $c_g(x_i)$ from the geometry basis grids and geometry coefficient grid via trilinear interpolation. By element-wise multiplying the geometry basis factor with the geometry coefficient factor of the sampled point, we obtain the geometry feature $f_g(x_i)$ of $x_i$. For the appearance feature $f_a(x_i)$ of sampled point $x_i$, we obtain it similarly. Then, we input the geometry feature of the sampled point into the geometry decoder $\mathbf{MLP}_g$ to decode the truncated signed distance (TSDF) value $s(x_i)$ of $x_i$.
\begin{eqnarray}
s(x_{i})=\mathbf{MLP}_g(f_g(x_i))
\end{eqnarray}
In order to allocate sampled point weight for rendering depth and color, we adopt a method similar to \cite{ESLAM} to convert the TSDF value into weight for rendering along the ray.
\begin{eqnarray}
\sigma(x_{i})=1-\exp \left(-\beta \cdot \operatorname{sigmod}\left(-s(x_{i}) \cdot \beta\right)\right) 
\end{eqnarray}
\begin{eqnarray}
w_{i}=\sigma(x_{i}) \prod_{j=1}^{i-1}\left(1-\sigma(x_{j})\right)
\end{eqnarray}
where $\sigma(x_{i})$ is the volume density of $x_i$ and $\beta$ is a hyperparameter. Therefore, the rendered depth of the ray is
\begin{eqnarray}
\hat{D}(r)=\sum_{i=1}^{N} w_{i} z_{i}
\end{eqnarray}

\subsection{Feature Integration Rendering}\label{sec:3.2}

Previous neural implicit dense visual SLAM methods \cite{iMAP, NICE-SLAM, ESLAM, Co-SLAM, Vox-Fusion, point-slam} obtain rendered pixel color by querying the raw color of sampled points and relying on volume rendering by marching camera ray. However, this rendering approach is relatively inefficient for implementing real-time neural SLAM system. In our work, instead of decoding the appearance features of sampled points to raw colors, we input the appearance feature of the entire ray into the color decoder to obtain the final rendered pixel color. Compared to \cite{fr}, which uses a large MLP to ensure color rendering quality, our method requires only a shallow MLP to achieve high-fidelity color reconstruction, thereby reducing the number of parameters that requires optimization during backpropagation. This further balances the quality and efficiency of the neural implicit SLAM system. Specifically, after obtaining the appearance features and weights of all sampled points along the ray, we approximate the overall appearance feature of the ray by performing a weighted summation of the appearance features of all sampled points along the ray: 
\begin{eqnarray}
f_a(r)=\sum_{i=1}^{N} w_{i}f_a(x_{i})
\end{eqnarray}
The ray feature $f_a(r)$ are decoded into rendered pixel color by the color decoder $\mathbf{MLP}_a$.
\begin{eqnarray}
\hat{C}(r)=\mathbf{MLP}_a(f_a(r))
\end{eqnarray}

\subsection{Tracking and Mapping}\label{sec:3.3}
\subsubsection{Loss Functions}
To optimize scene representation and camera poses, we apply color loss, depth loss, free-space loss and SDF loss. The color loss is the $\mathcal{L}_{2}$ loss between the rendered pixel color $\hat{C}(r)$ and the ground truth pixel color $C(r)$: 
\begin{eqnarray}
\mathcal{L}_c = \frac{1}{\left | R \right |} \sum_{r \in R} (C(r) - \hat{C}(r)) ^2
\end{eqnarray}
Similarly, the depth loss is the $\mathcal{L}_{2}$ loss between the ground truth depth $D(r)$ and the rendered depth $\hat{D}(r)$ of the pixel.
\begin{eqnarray}
\mathcal{L}_d = \frac{1}{\left | R_{vd} \right | }  \sum_{r \in R_{vd}} (D(r) - \hat{D}(r)) ^2
\end{eqnarray}
where $R$ is a set of pixels, $R_{vd}$ denotes camera rays with valid depth in $R$. To achieve more accurate surface reconstruction, We employ depth sensor measurement to approximate the SDF loss. When sampled point is far from the truncation region, we apply the free-space loss $L_{fs}$ to enforce consistency between the TSDF value and the truncation distance. When the sampled point within the truncation region $tr$, i.e., $\left|z_{i}-D(r)\right|<tr$, we apply SDF loss $L_{sdf}$ to learn the surface geometry within the truncated region.

\begin{eqnarray}
\mathcal{L}_{fs} = \frac{1}{|R_{vd}|} \sum_{r \in R_{vd}} \frac{1}{|P^{fs}_r|} \sum_{x_i \in P^{fs}_r}^{} (s(x_i)- tr)^2
\end{eqnarray}
\begin{eqnarray}
\mathcal{L}_{sdf} = \frac{1}{|R_{vd}|} \sum_{r \in R_{vd}} \frac{1}{|P^{t}_r|} \sum_{x_i \in P^{t}_r}^{} (s(x_i) + z_i- D(r))^2
\end{eqnarray}
Where $P^{fs}_r$ denotes the sampled points along the ray located between the camera origin and the surface truncation region of the depth sensor measurement, while $P^{t}_r$ denotes the sampled points along the ray within the truncation region.

The global loss function we use is as follows:
\begin{eqnarray}
\mathcal{L} = \lambda_{c}\mathcal{L}_c + \lambda_{d}\mathcal{L}_d + \lambda_{fs}\mathcal{L}_{fs} + \lambda_{sdf}\mathcal{L}_{sdf}
\end{eqnarray}
where $\lambda_{c}$, $\lambda_{d}$, $\lambda_{fs}$ and $\lambda_{sdf}$ are weighting coefficients. We employ the same global loss function in both tracking and mapping, but with different weighting coefficients set. Furthermore, following \cite{ESLAM}, we use different SDF loss coefficients for sampling points at the center and the tail of the truncation region.

\begin{figure*}
	\centering
	\includegraphics[width=12cm]{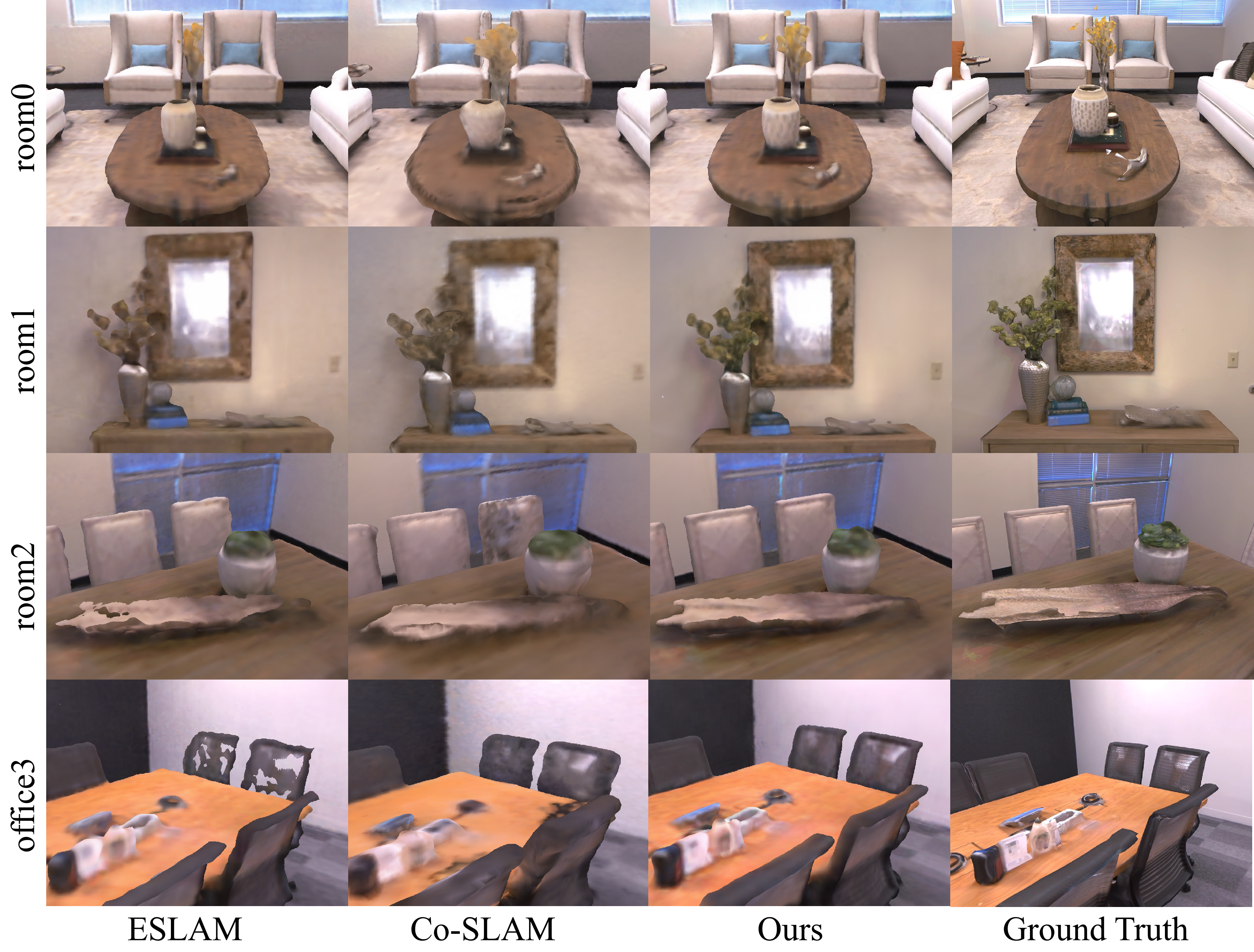}
	\caption{Comparison of qualitative results of reconstruction on Replica dataset \cite{replica}.} 
   \label{fig:2}
\end{figure*}

\subsubsection{Tracking}\label{subsec2}
During tracking, we obtain a copy of the factor grids and MLP from the mapping process and keep the parameters in them fixed, and only optimize the transformation matrix $\mathbf{T}_{i}\in SE(3)$ in the world coordinate for frame $i$. At the start of tracking process, We apply constant speed assumption to initialize the camera pose for frame $i$: 
\begin{eqnarray}
\mathbf{T}_{i}=\mathbf{T}_{i-1} \mathbf{T}_{i-2}^{-1} \mathbf{T}_{i-1}
\end{eqnarray}
For current frame, $\left | R_t \right |$ pixels are randomly selected to participate in iterative optimization. During this process, we only update the camera pose, while the factor grids and decoders remain fixed.

\subsubsection{Mapping}\label{subsec2}
For the first input frame, we fix the camera pose and only optimize the scene representation. For subsequent input frames, we jointly optimize the scene representation and camera poses every $k$ frames similar to \cite{NICE-SLAM, ESLAM, Co-SLAM}. During jointly optimization, we optimize a total of $|W|$ frames, which include $|W-2|$ keyframes selected from the keyframe list that are co-viewed with the current frame, as well as the last keyframe and the current frame. After obtaining the RGB-D information of these $|W|$ frames, we randomly select $\left | R_m \right |$ pixels for iterative optimization.

\section{Experiments}\label{sec4}
\subsection{Experimental Setup}\label{subsec2}
\subsubsection{Datasets}\label{subsec2}
We evaluate our method on the Replica \cite{replica}, ScanNet \cite{scannet}, and TUM-RGBD \cite{tum-rgbd} datasets. In this paper, we select 8 synthetic scenes from Replica \cite{replica} and 6 real-world scenes from ScanNet \cite{scannet} to evaluate localization and reconstruction, and 3 real-word scenes from TUM-RGBD \cite{tum-rgbd} to evaluate localization. For tracking evaluation, the ground-truth camera poses of ScanNet \cite{scannet} are obtained with BundleFusion \cite{bundlefusion}.

\begin{figure*}
	\centering
	\includegraphics[width=12cm]{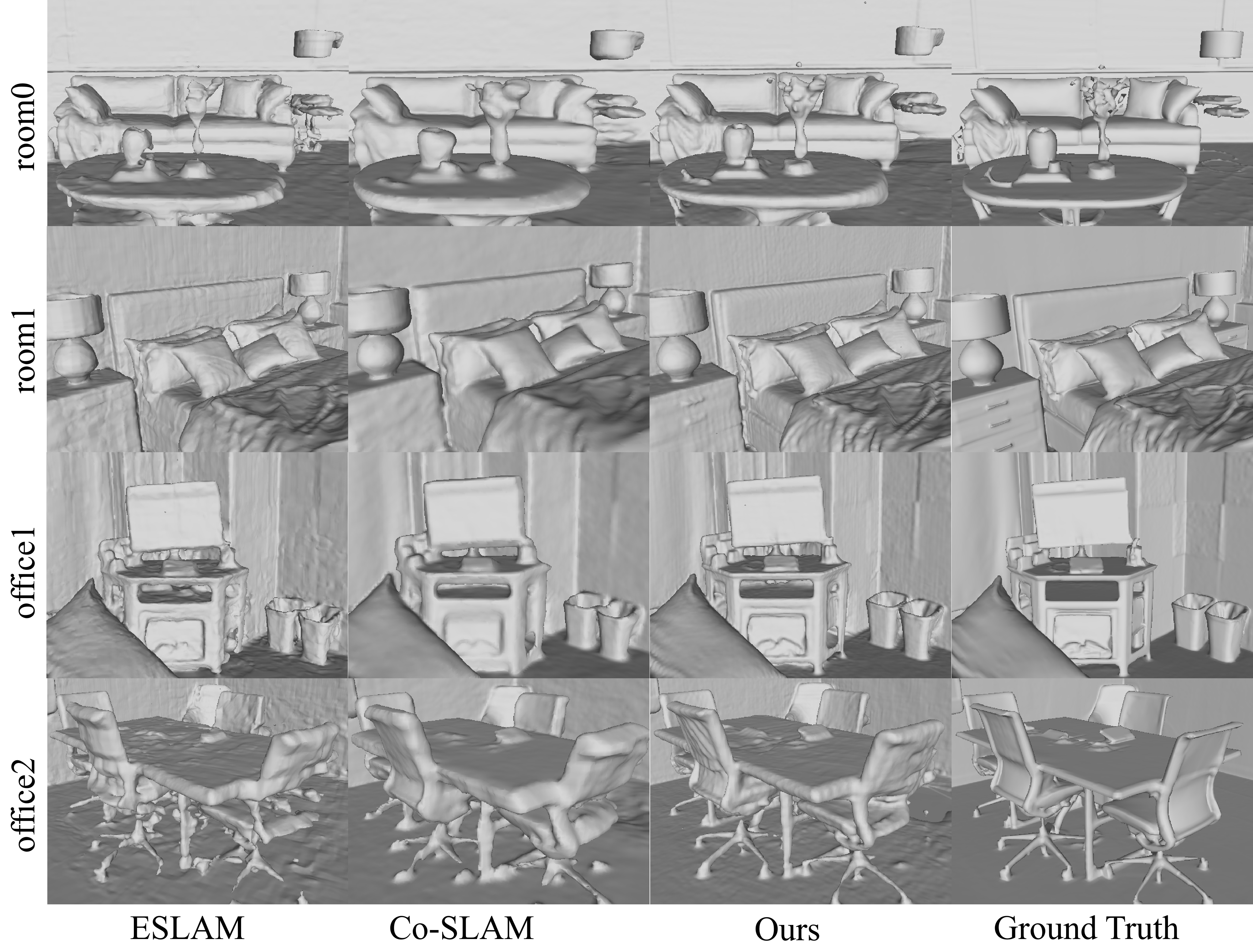}
	\caption{Comparison of qualitative results of reconstruction on Replica dataset \cite{replica}. We visualize untextured meshes.} 
   \label{fig:3}
\end{figure*}

\subsubsection{Baseline}\label{subsec2}
We compare our method to existing state-of-the-art neural implicit SLAM methods: iMAP \cite{iMAP}, NICE-SLAM\cite{NICE-SLAM}, Vox-Fusion \cite{Vox-Fusion}, ESLAM \cite{ESLAM}, Co-SLAM \cite{Co-SLAM}, Point-SLAM \cite{point-slam} and GS-SLAM \cite{GS-SLAM}. Note that iMAP* and Vox-Fusion* denote the results we reproduced using the corresponding open-source codes.

\begin{table*}[ht]
\caption{Reconstruction results in 8 synthetic scenes of the Replica dataset \cite{replica}.} 
\label{tab:1}
\centering
\scalebox{0.79}{
\setlength{\tabcolsep}{1mm}{
\begin{tabular}{c|cccccccccc}
\hline
{ Method}                      & Metric                          &  room0 & room1 & room2                            & office0        & office1        & office2        & office3        & office4        & Avg.           \\ \hline
                                                   & { Acc.↓}             & { 5.75}          & { 5.44}          & { 6.32}          & 7.58           & 10.25          & 8.91           & 6.89           & 5.34           & 7.06           \\
                                                   & { Comp.↓}            & { 5.96}          & { 5.38}          & { 5.21}          & 5.16           & 5.49           & 6.04           & 5.75           & 6.57           & 5.69           \\
                                                   & { Comp. Ratio (\%)↑} & { 67.13}         & { 68.91}         & { 71.69}         & 70.14          & 73.47          & 63.94          & 67.68          & 62.30          & 68.16          \\
\multirow{-4}{*}{iMAP* \cite{iMAP}}                             & { Depth L1↓}         & { 5.55}          & { 5.47}          & { 6.93}          & 7.63           & 8.13           & 10.61          & 9.66           & 8.44           & 7.8            \\ \hline
{ }                            & { Acc.↓}             & { 1.47}          & { 1.21}          & { 1.66}          & 1.28           & 1.02           & 1.54           & 1.95           & 1.60           & 1.47           \\
{ }                            & { Comp.↓}   & { 1.51} & { 1.27} & { 1.65} & 1.74           & 1.04           & 1.62           & 2.57           & 1.67           & 1.63           \\
{ }                            & Comp. Ratio (\%)↑                        & 98.16                                & 98.71                                & 96.42                                & 95.98          & 98.83          & 97.19          & 92.05          & 97.34          & 96.83          \\
\multirow{-4}{*}{{ NICE-SLAM \cite{NICE-SLAM}}} & Depth L1↓                                & 3.38                                 & 3.03                                 & 3.76                                 & 2.62           & 2.31           & 4.12           & 8.19           & 2.73           & 3.77           \\ \hline
                                                   & Acc.↓                                    & 1.08                                 & 1.05                                 & 1.21                                 & 1.41           & 0.82           & 1.31           & 1.34           & 1.31           & 1.19           \\
                                                   & Comp.↓                                   & 1.07                                 & 1.97                                 & 1.62                                 & 1.58           & 0.84           & 1.37           & 1.37           & 1.44           & 1.41           \\
                                                   & Comp. Ratio (\%)↑                        & 99.46                                & 94.76                                & 96.37                                & 95.80          & 99.12          & 98.20          & 97.55          & 97.32          & 97.32          \\
\multirow{-4}{*}{Vox-Fusion* \cite{Vox-Fusion}}                       & Depth L1↓                                & 1.62                                 & 10.43                                & 3.06                                 & 4.12           & 2.05           & 2.85           & 3.11           & 4.22           & 3.93           \\ \hline
                                                   & Acc.↓                                    & 1.11                                 & 1.14                                 & 1.17                                 & 0.99           & 0.76           & 1.36           & 1.44           & 1.24           & 1.15           \\
                                                   & Comp.↓                                   & 1.06                                 & 1.37                                 & 1.14                                 & 0.92           & 0.78           & 1.33           & 1.36           & 1.16           & 1.14           \\
                                                   & Comp. Ratio (\%)↑                        & 99.62                                & 97.49                                & 97.86                                & 99.07          & 99.25          & 98.81          & 98.48          & 98.96          & 98.69          \\
\multirow{-4}{*}{Co-SLAM \cite{Co-SLAM}}                          & Depth L1↓                                & 1.54                                 & 6.41                                 & 3.05                                 & 1.66           & 1.68           & 2.71           & 2.55           & 1.82           & 2.68           \\ \hline
                                                   & Acc.↓                                    & 1.07                                 & 0.85                                 & 0.93                                 & 0.85           & 0.83           & 1.02           & 1.21           & \textbf{0.97}  & 0.97           \\
                                                   & Comp.↓                                   & 1.12                                 & 0.88                                 & 1.05                                 & 0.96           & 0.81           & 1.09           & 1.42           & \textbf{1.05}  & 1.05           \\
                                                   & Comp. Ratio (\%)↑                        & 99.06                                & 99.64                                & 98.84                                & 98.34          & 98.85          & 98.60          & 96.80          & 98.60          & 98.60          \\
\multirow{-4}{*}{ESLAM \cite{ESLAM}}                            & Depth L1↓                                & 0.97                                 & 1.07                                 & 1.28                                 & 0.86           & 1.26           & 1.71           & 1.43           & 1.18           & 1.18           \\ \hline
                                                   & Acc.↓                                    & -                                  & -                                  & -                                  & -            & -            & -            & -            & -            & -            \\
                                                   & Comp.↓                                   & -                                  & -                                  & -                                  & -            & -            & -            & -            & -            & -            \\
                                                   & Comp. Ratio (\%)↑                        & -                                  & -                                  & -                                  & -            & -            & -            & -            & -            & -            \\
\multirow{-4}{*}{GS-SLAM \cite{GS-SLAM}}                          & Depth L1↓                                & 1.31                                 & 0.82                                 & 1.26                                 & 0.81           & 0.96           & 1.41           & 1.53           & 1.08           & 1.16           \\ \hline
                                                   & Acc.↓                                    & \textbf{0.98}                        & \textbf{0.76}                        & \textbf{0.84}                        & \textbf{0.76}  & \textbf{0.59}  & \textbf{0.90}  & \textbf{1.06}  & 1.0            & \textbf{0.86}  \\
                                                   & Comp.↓                                   & \textbf{0.99}                        & \textbf{0.78}                        & \textbf{0.94}                        & \textbf{0.79}  & \textbf{0.66}  & \textbf{0.94}  & \textbf{1.11}  & 1.07           & \textbf{0.91}  \\
                                                   & Comp. Ratio (\%)↑                        & \textbf{99.65}                       & \textbf{99.76}                       & \textbf{99.13}                       & \textbf{99.50} & \textbf{99.41} & \textbf{99.21} & \textbf{98.95} & \textbf{99.23} & \textbf{99.36} \\
\multirow{-4}{*}{Ours}                             & Depth L1↓                                & \textbf{0.84}                        & \textbf{0.76}                        & \textbf{1.04}                        & \textbf{0.68}  & \textbf{0.90}  & \textbf{1.24}  & \textbf{1.07}  & \textbf{0.61}  & \textbf{0.89}  \\ \hline
\end{tabular}

}
}
\end{table*}

\subsubsection{Evaluation Metrics}\label{subsec2}
We evaluate the reconstructed mesh using 3D metrics Accuracy (cm), Completion (cm), and Completion Ratio (\%) and a 2D metric Depth L1 (cm). Accuracy (cm) is then defined as the average distance between points on the reconstructed mesh and their nearest points on the ground truth mesh. Completion (cm) is similarly defined as the average distance between sampled points on the ground-truth mesh and their nearest sampled points on the reconstructed mesh. Completion Ratio (\%) denotes the percentage of points in the reconstructed mesh with a completion distance under than 5 cm. For Depth L1 (cm), following \cite{ESLAM}, we render depth in both the reconstructed and ground truth meshes for 1000 synthetic views. These views are uniformly sampled within the mesh, and views not observed by the input frames are rejected. For the evaluation of camera localization, we adopt ATE RMSE (cm) \cite{tum-rgbd}.

\begin{figure*}
	\centering
	\includegraphics[width=12cm]{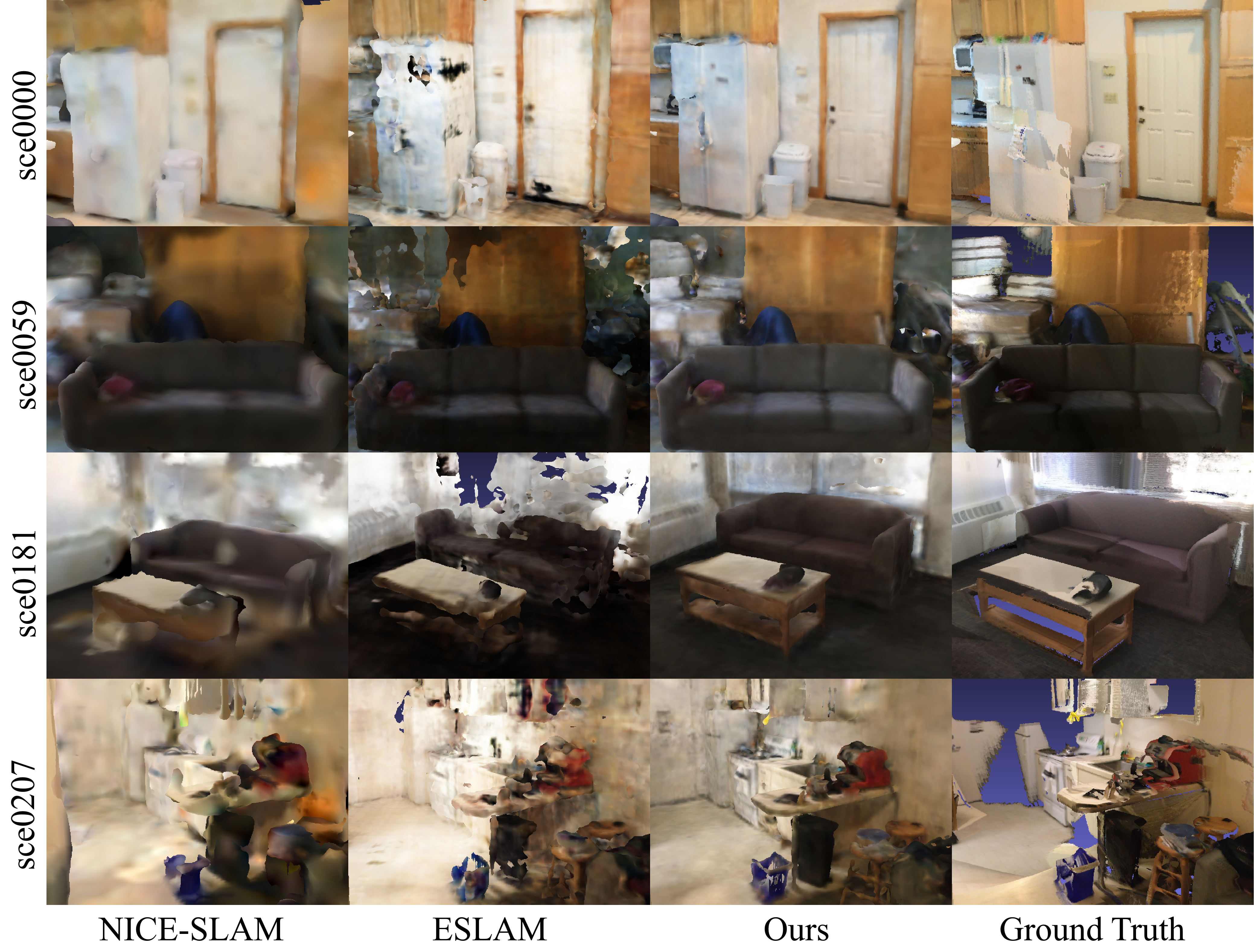}
	\caption{Comparison of qualitative results of reconstruction on ScanNet \cite{scannet}.} 
   \label{fig:4}
\end{figure*}

\subsubsection{Experimental Details.}\label{subsec2}
We uniformly set the truncation distance $tr$ to 6 cm, $\beta$ to 10. We employ a total of 6 levels of grids with linearly increasing resolution as our basis grids. We set resolutions similar to \cite{factor-field}: $[32, 128]^{T} \cdot \frac{\min (a-b)}{1024}$, where $a$, $b$ are scene bounding box. The channels for each level of basis grids are $[4, 4, 4, 2, 2, 2]^{T}$, respectively. The coefficient grid has single level, with a resolution of 32. Both geometry and appearance feature vectors have 18 channels. The geometry decoder is a single-layer MLP with 64 channels in the hidden layer, color decoder use a two-layer MLP with 128 channels. We default use $R_t$ = 2000 pixels for tracking and $R_m$ = 4000 pixels from every 4th frame for mapping. For the Replica dataset \cite{replica}, we perform 8 iterations for tracking and 15 iterations for mapping. On the ScanNet \cite{scannet} and TUM-RGBD \cite{tum-rgbd} datasets, we perform 15 iterations for tracking and 30 iterations for mapping. Quantitative results for tracking and mapping are averages of 5 runs.

\begin{table*}[ht]
\caption{Camera localization results (ATE RMSE [cm] ↓) in 8 synthetic scenes of Replica dataset \cite{replica}.} 
\label{tab:2}
\centering
\scalebox{0.79}{
\setlength{\tabcolsep}{2.5mm}{
\begin{tabular}{c|ccccccccc}
\hline
Method     & room0           & room1           & room2           & office0          & office1          & office2          & office3          & office4          & Avg.          \\ \hline
iMAP* \cite{iMAP}       & 3.88          & 3.01          & 2.43          & 2.67          & 1.07          & 4.68          & 4.83          & 2.48          & 3.13          \\
NICE-SLAM \cite{NICE-SLAM}   & 1.76          & 1.97          & 2.2           & 1.44          & 0.92          & 1.43          & 2.56          & 1.55          & 1.73          \\
VoxFusion* \cite{Vox-Fusion}  & 0.73          & 1.1           & 1.1           & 7.4           & 1.26          & 1.87          & 0.93          & 1.49          & 1.98          \\
CoSLAM \cite{Co-SLAM}     & 0.82          & 2.03          & 1.34          & 0.6           & 0.65          & 2.02          & 1.37          & 0.88          & 1.21          \\
ESLAM \cite{ESLAM}      & 0.71          & 0.7           & 0.52          & 0.57          & 0.55          & 0.58          & 0.72          & 0.63          & 0.63          \\
Point-SLAM \cite{point-slam} & 0.61          & 0.41          & 0.37          & 0.38          & 0.48          & 0.54          & 0.69          & 0.72          & 0.52          \\
GS-SLAM \cite{GS-SLAM}    & \textbf{0.48}       & 0.53          & \textbf{0.33} & 0.52          & 0.41          & 0.59          & 0.46          & 0.70          & 0.50          \\ \hline
Ours       & \textbf{0.48} & \textbf{0.37} & 0.35          & \textbf{0.32} & \textbf{0.24} & \textbf{0.43} & \textbf{0.38} & \textbf{0.35} & \textbf{0.37} \\ \hline
\end{tabular}

}
}
\end{table*}

\begin{table*}[ht]
\caption{Camera localization results (ATE RMSE [cm] ↓) in 6 real-world scenes of ScanNet dataset \cite{scannet}.} 
\label{tab:3}
\centering
\scalebox{0.79}{
\setlength{\tabcolsep}{2.5mm}{
\begin{tabular}{c|ccccccc}
\hline
{Scene ID} & {0000} & {0059} & {0106} & {0169} & {0181} & {0207} & {Avg.}                   \\ \hline
iMAP* \cite{iMAP}                    & 32.2                     & 17.3                     & 12.0                     & 17.4                     & 27.9                     & 12.7                     & 19.42                 \\
NICE-SLAM \cite{NICE-SLAM}                & 13.3                     & 12.8                     & 7.8                      & 13.2                     & 13.9                     & 6.2                      & 11.2                  \\
VoxFusion* \cite{Vox-Fusion}               & 11.6                     & 26.3                     & 9.1                      & 32.3                     & 22.1                     & 7.4                      & 18.13                 \\
CoSLAM \cite{Co-SLAM}                  & 7.9                      & 12.6                     & 9.5                      & 6.6                      & 12.9                     & 7.1                      & 9.43                  \\
ESLAM \cite{ESLAM}                   & 7.3                      & 8.5                      & 7.5                      & 6.5                      & \textbf{9.0}             & 5.7                      & 7.4                   \\
Point-SLAM \cite{point-slam}              & 10.24                    & 7.81                     & 8.65                     & 22.16                    & 14.77                    & 9.54                     & 12.19                 \\ \hline
Ours                    & \textbf{6.8}             & \textbf{7.8}             & \textbf{7.2}             & \textbf{5.6}             & 10.3                     & \textbf{4.2}             & \textbf{6.98}         \\ \hline
\end{tabular}

}
}
\end{table*}

\begin{table*}[ht]
\caption{Camera localization results (ATE RMSE [cm] ↓) in 3 real-world scenes of TUM-RGBD \cite{tum-rgbd}.} 
\label{tab:4}
\centering
\scalebox{0.79}{
\setlength{\tabcolsep}{2.5mm}{
\begin{tabular}{c|cccc}
\hline
Method     & fr1/desk      & fr2/xyz       & fr3/office    & Avg.          \\ \hline
iMAP* \cite{iMAP}       & 5.9           & 2.2           & 7.6           & 5.23          \\
NICE-SLAM \cite{NICE-SLAM}   & 2.72          & 31            & 15.2          & 16.31         \\
VoxFusion* \cite{Vox-Fusion}  & 3.2           & 1.6           & 25.4          & 10.06         \\
CoSLAM \cite{Co-SLAM}     & 2.88          & 1.85          & 2.91          & 2.55          \\
ESLAM \cite{ESLAM}      & 2.47          & \textbf{1.11} & \textbf{2.42} & 2.00             \\
Point-SLAM \cite{point-slam} & 4.34          & 1.31          & 3.48          & 3.04          \\
GS-SLAM \cite{GS-SLAM}    & 3.3           & 1.3           & 6.6           & 3.7           \\ \hline
Ours       & \textbf{2.11} & 1.32          & 2.48          & \textbf{1.97} \\ \hline
\end{tabular}
}
}
\end{table*}

\subsection{Reconstruction Evaluation}\label{subsec2}
\begin{figure*}
	\centering
	\includegraphics[width=12cm]{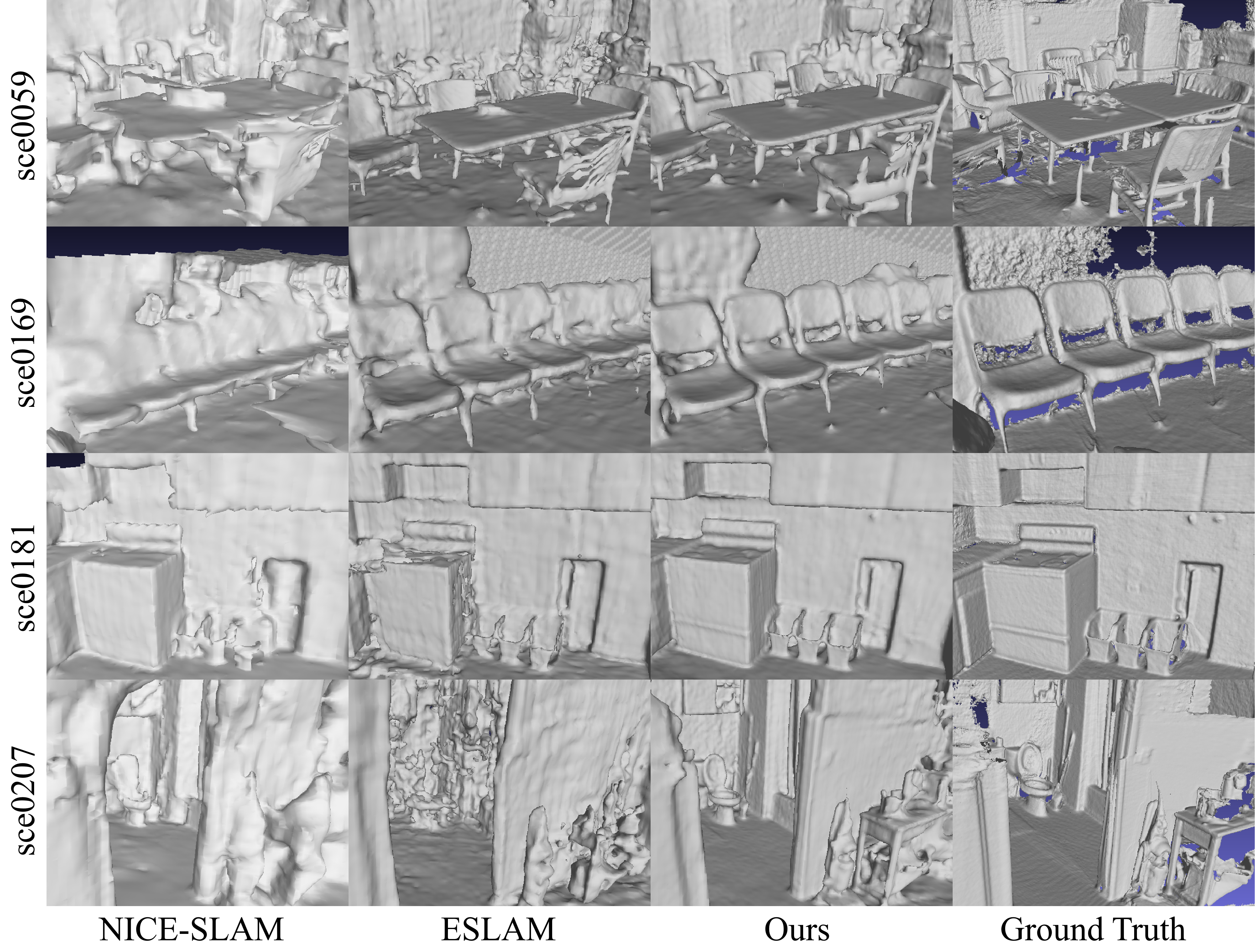}
	\caption{Qualitative reconstruction on ScanNet dataset \cite{scannet}. We visualize untextured meshes.} 
   \label{fig:5}
\end{figure*}

We show qualitative comparisons of the Replica \cite{replica} and ScanNet \cite{scannet} datasets in Fig.\ref{fig:2}, Fig.\ref{fig:3}, Fig.\ref{fig:4} and Fig.\ref{fig:5}, and report quantitative comparisons of the Replica dataset \cite{replica} in Tab.\ref{tab:1}. We use meshlab \cite{meshlab} to visualize all meshes. Due to the incomplete ground truth mesh of the ScanNet dataset \cite{scannet}, we only provide a qualitative analysis of geometric reconstructions for this dataset, similar to previous work \cite{iMAP, NICE-SLAM, ESLAM, Co-SLAM, point-slam}. Our method outperforms the baselines on average, and compared to other baseline methods, it exhibits superior performance in recovering fine details of the scenes. Our method achieves high-fidelity scene reconstruction even in real-world scenes with noise, and is able to complete unobserved areas of the input frame.

\subsection{Camera Localization Evaluation}\label{subsec2}
\begin{figure}
	\centering
	\includegraphics[width=12cm]{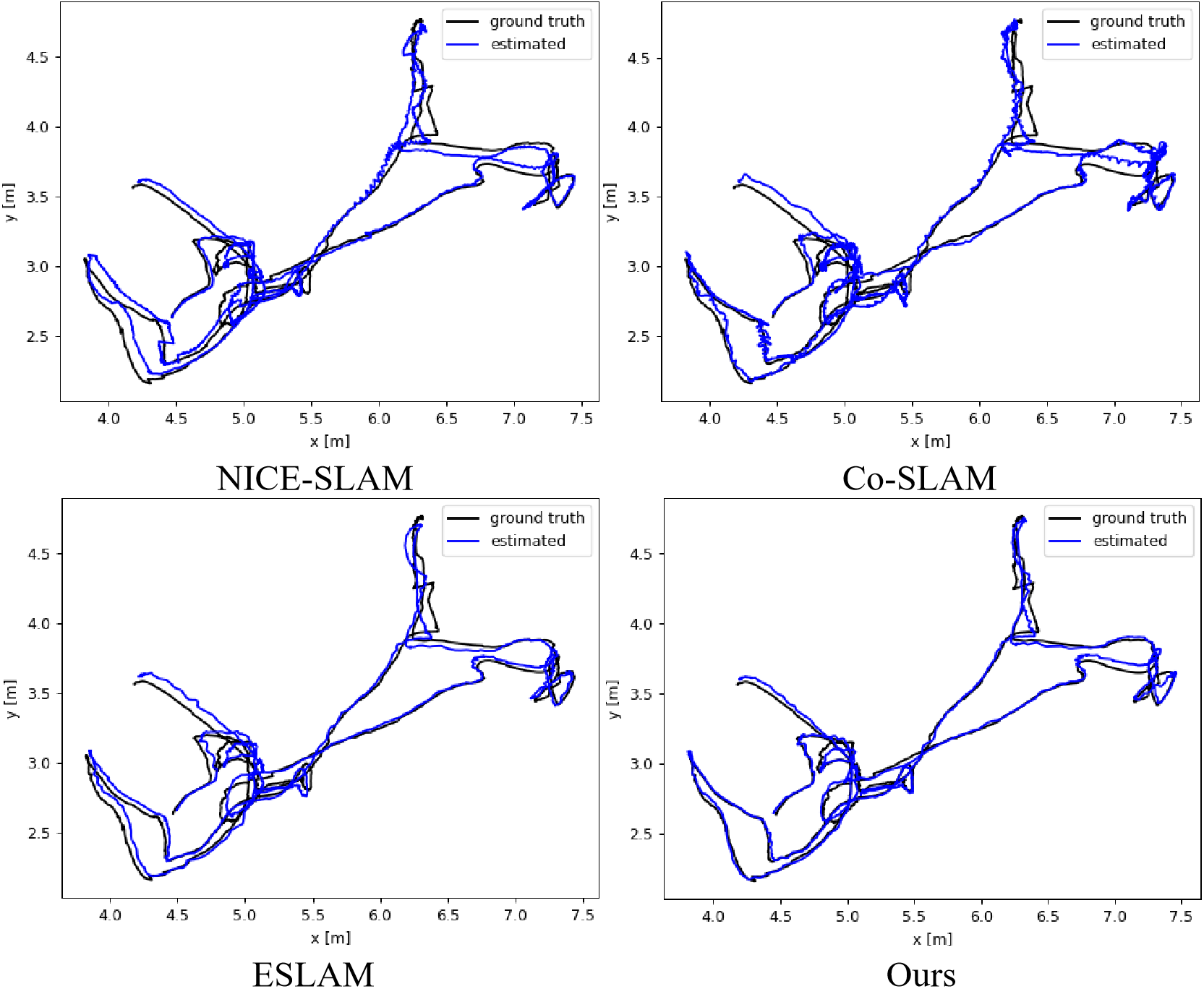}
	\caption{Trajectories of NICE-SLAM \cite{NICE-SLAM}, ESLAM \cite{ESLAM}, Co-SLAM \cite{Co-SLAM} and our method on scene0207 of ScanNet dataset \cite{scannet}.} 
   \label{fig:6}
\end{figure}

We report quantitative comparisons on the three datasets in Tab.\ref{tab:2}, Tab.\ref{tab:3} and Tab.\ref{tab:4}. In addition, Fig.\ref{fig:6} shows a qualitative comparison of camera trajectories on the scene0207 scene of the ScanNet dataset \cite{scannet}. Our method has more robust tracking performance in both synthetic and real-world datasets, and significantly reduces the impact of trajectory drift. This is due to the fact that our method has a more accurate scene representation and is able to guide the tracking process to achieve higher camera localization accuracy.

\begin{figure}
	\centering
	\includegraphics[width=12cm]{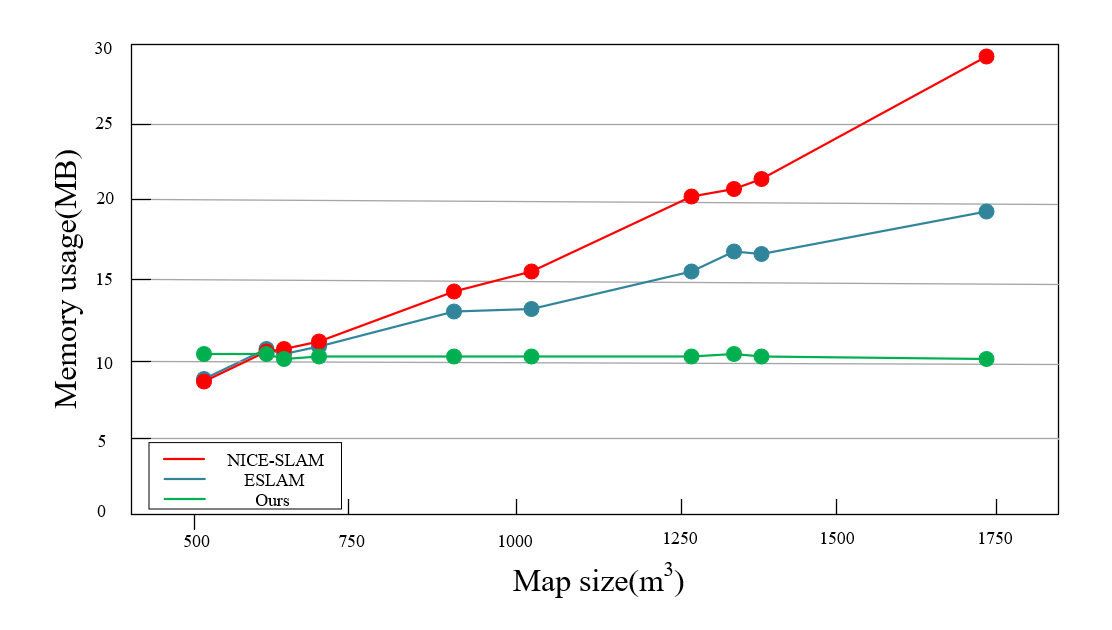}
	\caption{Comparison of the relationship between map size and memory usage changes. We select some scenes from the Replica \cite{replica} and ScanNet \cite{scannet} datasets for evaluation.} 
   \label{fig:7}
\end{figure}

\begin{table*}[ht]
\caption{Runtime and memory usage comparison on Replica\cite{replica} room0.} 
\label{tab:5}
\centering
\scalebox{0.79}{
\setlength{\tabcolsep}{2.5mm}{
\begin{tabular}{c|cccc}
\hline
\multirow{2}{*}{Method}
                        & Tracking time & Mapping time & FPS           & Param.           \\
                        & (ms/it)↓      & (ms/it)↓     & (Hz)↑         & (MB)↓           \\ \hline
iMAP* \cite{iMAP}             & 34.63         & 20.15        & 0.18          & \textbf{0.22}   \\
NICE-SLAM \cite{NICE-SLAM}         & 7.48          & 30.59        & 0.71          & 11.56           \\
Vox-Fusion* \cite{Vox-Fusion}      & 11.2          & 52.7         & 1.67          & 1.19            \\
Co-SLAM \cite{Co-SLAM}           & \textbf{6.15}          & 14.33        & \textbf{10.5}          & 0.26            \\
ESLAM \cite{ESLAM}             & 7.11          & 20.32        & 5.6           & 6.79            \\
Point-SLAM \cite{point-slam}   & 12.23         & 35.21        & 0.29          & 27.23           \\ 
GS-SLAM \cite{GS-SLAM}           & 11.90         & 12.80        & 8.34            & 198.04 \\ \hline
Ours                    & 6.47          & \textbf{10.14}        & 9.8           & 10.15           \\ \hline
\end{tabular}
}
}
\end{table*}

\subsection{Runtime and Memory Usage Analysis}\label{subsec2}
We report the runtime and memory usage on the room0 scene of the Replica dataset \cite{replica} in Tab.\ref{tab:5}. The runtimes were analyzed on a single NVIDIA RTX 3090, while the runtime of GS-SLAM \cite{GS-SLAM} are reported by the authors using a single NVIDIA RTX 4090, since they have not released executable code. FPS (Hz) is calculated based on the total runtime of the system, and the number of parameters is the sum of the model size of the scene representation and decoders. Our method has the fastest iteration time during the mapping process. In addition, we discuss the effect of scene map size on memory usage in Fig.\ref{fig:7}. For memory usage, both NICE-SLAM \cite{NICE-SLAM} and ESLAM \cite{ESLAM} are sensitive to map size, while our method is not affected by map size, which makes our method more suitable for large-scale scenes.

\begin{figure}
	\centering
	\includegraphics[width=13cm]{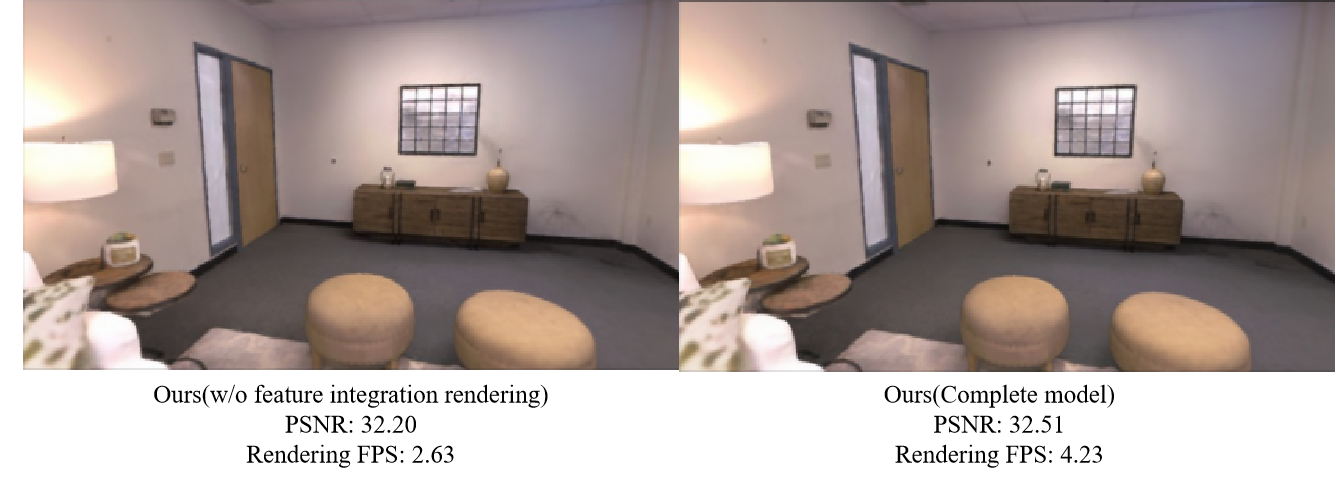}
	\caption{Ablation results of feature integration rendering on Replica \cite{replica} room0.} 
   \label{fig:8}
\end{figure}

\begin{table*}[ht]
\caption{Ablation results of reconstruction in 8 synthetic scenes of Replica dataset \cite{replica}.} 
\label{tab:6}
\scalebox{0.79}{
\setlength{\tabcolsep}{0.75mm}{
\begin{tabular}{c|cccccccccc}
\hline
Method                & Metric            & room0          & room1          & room2          & office0        & office1        & office2        & office3        & office4        & Avg.           \\ \hline
                      & Acc.↓             & 1              & 0.82           & 0.86           & 0.76           & 0.66           & 0.98           & 1.13           & 1.08           & 0.91           \\
w/o separate          & Comp.↓            & 1.04           & 0.85           & 0.96           & 0.8            & 0.72           & 1              & 1.17           & 1.12           & 0.94           \\
factor grids          & Comp. Ratio (\%)↑ & 99.47          & 99.62          & 99.14          & 99.54          & 99.31          & 99.35          & 98.90          & 99.09          & 99.3           \\
                      & Depth L1↓         & 0.92           & 1.13           & 1.22           & 0.75           & 1.20           & 1.17           & 1.53           & 1.31           & 1.22           \\ \hline
                      & Acc.↓             & 1.11           & 0.97           & 1.03           & 1.05           & 1.06           & 1.0            & 1.18           & 1.12           & 1.07           \\
w/o multi-level       & Comp.↓            & 1.04           & 1              & 1.24           & 0.93           & 0.98           & 0.98           & 1.19           & 1.16           & 1.06           \\
basis grids           & Comp. Ratio (\%)↑ & 99.63          & 99.49          & 97.46          & 99.58          & 98.62          & 99.52          & 98.99          & 98.92          & 99.03          \\
                      & Depth L1↓         & 0.91           & 1.65           & 2.75           & 1.12           & 1.76           & 1.66           & 1.67           & 1.29           & 1.60           \\ \hline
                      & Acc.↓             & 0.98           & 0.79           & 0.86           & 0.79           & 0.73           & 0.93           & 1.08           & 1.03           & 0.90           \\
w/o feature           & Comp.↓            & 1.01           & 0.81           & 0.98           & 0.82           & 0.71           & 0.96           & 1.14           & 1.09           & 0.94           \\
integration rendering & Comp. Ratio (\%)↑ & 99.51          & 99.73          & 98.94          & 99.55          & 99.32          & 99.41          & 98.84          & 99.28          & 99.32          \\
                      & Depth L1↓         & 0.82           & 0.89           & 1.22           & 0.7            & 1.01           & 1.57           & 1.23           & 0.77           & 1.03           \\ \hline
\multirow{4}{*}{Ours(Complete model)} & Acc.↓             & \textbf{0.98}  & \textbf{0.76}  & \textbf{0.84}  & \textbf{0.76}  & \textbf{0.59}  & \textbf{0.90}  & \textbf{1.06}  & \textbf{1.0}   & \textbf{0.86}  \\
                      & Comp.↓            & \textbf{0.99}  & \textbf{0.78}  & \textbf{0.94}  & \textbf{0.79}  & \textbf{0.66}  & \textbf{0.94}  & \textbf{1.11}  & \textbf{1.07}  & \textbf{0.91}  \\
                      & Comp. Ratio (\%)↑ & \textbf{99.65} & \textbf{99.76} & \textbf{99.13} & \textbf{99.50} & \textbf{99.41} & \textbf{99.21} & \textbf{98.95} & \textbf{99.23} & \textbf{99.36} \\
                      & Depth L1↓         & \textbf{0.84}  & \textbf{0.76}  & \textbf{1.04}  & \textbf{0.68}  & \textbf{0.90}  & \textbf{1.24}  & \textbf{1.07}  & \textbf{0.61}  & \textbf{0.89} 
\\ \hline
\end{tabular}
}
}
\end{table*}

\begin{table*}[ht]
\caption{Ablation results (ATE RMSE [cm] ↓) of camera localization in 8 synthetic scenes of Replica dataset \cite{replica}.} 
\label{tab:7}
\centering
\scalebox{0.79}{
\setlength{\tabcolsep}{2.3mm}{
\begin{tabular}{c|ccccccccc}
\hline
Method                & room0                          & room1                          & room2                          & office0                        & office1                        & office2                        & office3                        & office4                        & Avg.                           \\ \hline
w/o separate          & \multirow{2}{*}{0.62}          & \multirow{2}{*}{0.92}          & \multirow{2}{*}{0.59}          & \multirow{2}{*}{0.41}          & \multirow{2}{*}{0.54}          & \multirow{2}{*}{0.56}          & \multirow{2}{*}{0.56}          & \multirow{2}{*}{0.57}          & \multirow{2}{*}{0.59}          \\
factor grids          &                                &                                &                                &                                &                                &                                &                                &                                &                                \\ \hline
w/o multi-level       & \multirow{2}{*}{0.63}          & \multirow{2}{*}{1.67}          & \multirow{2}{*}{0.67}          & \multirow{2}{*}{0.92}          & \multirow{2}{*}{1.48}          & \multirow{2}{*}{0.56}          & \multirow{2}{*}{0.60}          & \multirow{2}{*}{0.86}          & \multirow{2}{*}{0.92}          \\
basis grids           &                                &                                &                                &                                &                                &                                &                                &                                &                                \\ \hline
w/o feature           & \multirow{2}{*}{0.54}          & \multirow{2}{*}{0.6}           & \multirow{2}{*}{0.42}          & \multirow{2}{*}{0.35}          & \multirow{2}{*}{0.24}          & \multirow{2}{*}{\textbf{0.41}} & \multirow{2}{*}{0.47}          & \multirow{2}{*}{0.36}          & \multirow{2}{*}{0.42}          \\
integration rendering &                                &                                &                                &                                &                                &                                &                                &                                &                                \\ \hline
\multirow{2}{*}{Ours(Complete model)} & \multirow{2}{*}{\textbf{0.48}} & \multirow{2}{*}{\textbf{0.37}} & \multirow{2}{*}{\textbf{0.35}} & \multirow{2}{*}{\textbf{0.32}} & \multirow{2}{*}{\textbf{0.24}} & \multirow{2}{*}{0.43}          & \multirow{2}{*}{\textbf{0.38}} & \multirow{2}{*}{\textbf{0.35}} & \multirow{2}{*}{\textbf{0.37}} \\
                      &                                &                                &                                &                                &                                &                                &                                &                                &                                \\ \hline
\end{tabular}
}
}
\end{table*}

\begin{table*}[ht]
\caption{Ablation results (ATE RMSE [cm] ↓) of camera localization in 6 real-world scenes of ScanNet \cite{scannet} dataset.} 
\label{tab:8}
\centering
\scalebox{0.79}{
\setlength{\tabcolsep}{2.5mm}{
\begin{tabular}{c|ccccccc}
\hline
Scene ID  & 0000         & 0059         & 0106         & 0169         & 0181 & 0207      & {Avg.}                      \\ \hline
w/o separate factor grids                 & 7.4          & 8.5          & 8.1          & 6.2          & 10.8 & 4.4          & 7.57                    \\
w/o multi-level basis grids                & 7.3          & 9.9          & 8.0          & 6.0          & 11.9 & 5.9          &   8.17                    \\
w/o feature integration rendering                 & 7.3          & 8.5          & 7.6          & 6.4          & 10.8 & 4.5          & 7.51                  \\ \hline
Ours(Complete model)                 & \textbf{6.8} & \textbf{7.8} & \textbf{7.2} & \textbf{5.6} & \textbf{10.3} & \textbf{4.2} & \textbf{6.98}         \\ \hline
\end{tabular}

}
}
\end{table*}

\subsection{Ablation Study}\label{subsec2}
In this section, we conduct a series of ablation experiments to validate our design choices. The experimental details are as follows: 1) We do not use separate factor grids, and instead employ a set of factor grids to represent both scene geometry and appearance. 2) We eliminate the multi-level basis grids setup, setting only a single level for basis grids. 3) We do not utilize feature integration rendering, instead continuing to employ traditional volume rendering.

Tab.\ref{tab:6}, Tab.\ref{tab:7} and Tab.\ref{tab:8} shows a quantitative comparison of the aforementioned three settings with our complete model in terms of localization and reconstruction on Replica \cite{replica} and ScanNet \cite{scannet} datasets. Our full model has higher accuracy and better completion than other setups. Fig.\ref{fig:8} shows the effect of using feature integration rendering on color rendering. Compared to our method without feature integration rendering, Our full model achieves a 60\% improvement in rendering speed while maintaining rendering quality.

\section{Conclusion}\label{sec5}
In this work, we present DF-SLAM, a high-fidelity neural implicit dense visual SLAM system. Extensive experiments demonstrate that the use of dictionary factors scene representation in neural SLAM system enables more detailed scene reconstruction and more efficient memory usage. Furthermore, we propose employing feature integration rendering instead of traditional volume rendering, which brings faster scene color reconstruction to our neural SLAM method with color quality comparable to using volume rendering methods. However, due to the utilization of feature integration rendering, our method is prone to artifacts in color rendering when dealing with input frames that have extreme motion blur. In the future, we will combine the deblurring module to solve this problem.


\begin{thebibliography}{10}
\providecommand{\url}[1]{#1}
\csname url@samestyle\endcsname
\providecommand{\newblock}{\relax}
\providecommand{\bibinfo}[2]{#2}
\providecommand{\BIBentrySTDinterwordspacing}{\spaceskip=0pt\relax}
\providecommand{\BIBentryALTinterwordstretchfactor}{4}
\providecommand{\BIBentryALTinterwordspacing}{\spaceskip=\fontdimen2\font plus
\BIBentryALTinterwordstretchfactor\fontdimen3\font minus \fontdimen4\font\relax}
\providecommand{\BIBforeignlanguage}[2]{{%
\expandafter\ifx\csname l@#1\endcsname\relax
\typeout{** WARNING: IEEEtran.bst: No hyphenation pattern has been}%
\typeout{** loaded for the language `#1'. Using the pattern for}%
\typeout{** the default language instead.}%
\else
\language=\csname l@#1\endcsname
\fi
#2}}
\providecommand{\BIBdecl}{\relax}
\BIBdecl

\bibitem{ORB-SLAM}
R.~Mur-Artal, J.~M.~M. Montiel, and J.~D. Tardos, ``Orb-slam: a versatile and accurate monocular slam system,'' \emph{IEEE transactions on robotics}, vol.~31, no.~5, pp. 1147--1163, 2015.

\bibitem{ORB-SLAM2}
R.~Mur-Artal and J.~D. Tard{\'o}s, ``Orb-slam2: An open-source slam system for monocular, stereo, and rgb-d cameras,'' \emph{IEEE transactions on robotics}, vol.~33, no.~5, pp. 1255--1262, 2017.

\bibitem{ORB-SLAM3}
C.~Campos, R.~Elvira, J.~J.~G. Rodr{\'\i}guez, J.~M. Montiel, and J.~D. Tard{\'o}s, ``Orb-slam3: An accurate open-source library for visual, visual--inertial, and multimap slam,'' \emph{IEEE Transactions on Robotics}, vol.~37, no.~6, pp. 1874--1890, 2021.

\bibitem{NeRF}
B.~Mildenhall, P.~P. Srinivasan, M.~Tancik, J.~T. Barron, R.~Ramamoorthi, and R.~Ng, ``Nerf: Representing scenes as neural radiance fields for view synthesis,'' \emph{Communications of the ACM}, vol.~65, no.~1, pp. 99--106, 2021.

\bibitem{iMAP}
E.~Sucar, S.~Liu, J.~Ortiz, and A.~J. Davison, ``imap: Implicit mapping and positioning in real-time,'' in \emph{Proceedings of the IEEE/CVF International Conference on Computer Vision}, 2021, pp. 6229--6238.

\bibitem{NICE-SLAM}
Z.~Zhu, S.~Peng, V.~Larsson, W.~Xu, H.~Bao, Z.~Cui, M.~R. Oswald, and M.~Pollefeys, ``Nice-slam: Neural implicit scalable encoding for slam,'' in \emph{Proceedings of the IEEE/CVF Conference on Computer Vision and Pattern Recognition}, 2022, pp. 12\,786--12\,796.

\bibitem{Vox-Fusion}
X.~Yang, H.~Li, H.~Zhai, Y.~Ming, Y.~Liu, and G.~Zhang, ``Vox-fusion: Dense tracking and mapping with voxel-based neural implicit representation,'' in \emph{2022 IEEE International Symposium on Mixed and Augmented Reality (ISMAR)}.\hskip 1em plus 0.5em minus 0.4em\relax IEEE, 2022, pp. 499--507.

\bibitem{ESLAM}
M.~M. Johari, C.~Carta, and F.~Fleuret, ``Eslam: Efficient dense slam system based on hybrid representation of signed distance fields,'' in \emph{Proceedings of the IEEE/CVF Conference on Computer Vision and Pattern Recognition}, 2023, pp. 17\,408--17\,419.

\bibitem{Co-SLAM}
H.~Wang, J.~Wang, and L.~Agapito, ``Co-slam: Joint coordinate and sparse parametric encodings for neural real-time slam,'' in \emph{Proceedings of the IEEE/CVF Conference on Computer Vision and Pattern Recognition}, 2023, pp. 13\,293--13\,302.

\bibitem{tensorf}
A.~Chen, Z.~Xu, A.~Geiger, J.~Yu, and H.~Su, ``Tensorf: Tensorial radiance fields,'' in \emph{European Conference on Computer Vision}.\hskip 1em plus 0.5em minus 0.4em\relax Springer, 2022, pp. 333--350.

\bibitem{factor-field}
A.~Chen, Z.~Xu, X.~Wei, S.~Tang, H.~Su, and A.~Geiger, ``Factor fields: A unified framework for neural fields and beyond,'' \emph{arXiv preprint arXiv:2302.01226}, 2023.

\bibitem{fr}
K.~Han, W.~Xiang, and L.~Yu, ``Volume feature rendering for fast neural radiance field reconstruction,'' \emph{arXiv preprint arXiv:2305.17916}, 2023.

\bibitem{PTAM}
G.~Klein and D.~Murray, ``Parallel tracking and mapping for small ar workspaces,'' in \emph{2007 6th IEEE and ACM international symposium on mixed and augmented reality}.\hskip 1em plus 0.5em minus 0.4em\relax IEEE, 2007, pp. 225--234.

\bibitem{DTAM}
R.~A. Newcombe, S.~J. Lovegrove, and A.~J. Davison, ``Dtam: Dense tracking and mapping in real-time,'' in \emph{2011 international conference on computer vision}.\hskip 1em plus 0.5em minus 0.4em\relax IEEE, 2011, pp. 2320--2327.

\bibitem{kinectfusion}
R.~A. Newcombe, S.~Izadi, O.~Hilliges, D.~Molyneaux, D.~Kim, A.~J. Davison, P.~Kohi, J.~Shotton, S.~Hodges, and A.~Fitzgibbon, ``Kinectfusion: Real-time dense surface mapping and tracking,'' in \emph{2011 10th IEEE international symposium on mixed and augmented reality}.\hskip 1em plus 0.5em minus 0.4em\relax Ieee, 2011, pp. 127--136.

\bibitem{BAD-SLAM}
T.~Schops, T.~Sattler, and M.~Pollefeys, ``Bad slam: Bundle adjusted direct rgb-d slam,'' in \emph{Proceedings of the IEEE/CVF Conference on Computer Vision and Pattern Recognition}, 2019, pp. 134--144.

\bibitem{droid-slam}
Z.~Teed and J.~Deng, ``Droid-slam: Deep visual slam for monocular, stereo, and rgb-d cameras,'' \emph{Advances in neural information processing systems}, vol.~34, pp. 16\,558--16\,569, 2021.

\bibitem{code-slam}
M.~Bloesch, J.~Czarnowski, R.~Clark, S.~Leutenegger, and A.~J. Davison, ``Codeslam—learning a compact, optimisable representation for dense visual slam,'' in \emph{Proceedings of the IEEE conference on computer vision and pattern recognition}, 2018, pp. 2560--2568.

\bibitem{node-slam}
E.~Sucar, K.~Wada, and A.~Davison, ``Nodeslam: Neural object descriptors for multi-view shape reconstruction,'' in \emph{2020 International Conference on 3D Vision (3DV)}.\hskip 1em plus 0.5em minus 0.4em\relax IEEE, 2020, pp. 949--958.

\bibitem{deep-slam}
R.~Li, S.~Wang, and D.~Gu, ``Deepslam: A robust monocular slam system with unsupervised deep learning,'' \emph{IEEE Transactions on Industrial Electronics}, vol.~68, no.~4, pp. 3577--3587, 2020.

\bibitem{neural}
T.~Takikawa, J.~Litalien, K.~Yin, K.~Kreis, C.~Loop, D.~Nowrouzezahrai, A.~Jacobson, M.~McGuire, and S.~Fidler, ``Neural geometric level of detail: Real-time rendering with implicit 3d shapes,'' in \emph{Proceedings of the IEEE/CVF Conference on Computer Vision and Pattern Recognition}, 2021, pp. 11\,358--11\,367.

\bibitem{plenoctrees}
A.~Yu, R.~Li, M.~Tancik, H.~Li, R.~Ng, and A.~Kanazawa, ``Plenoctrees for real-time rendering of neural radiance fields,'' in \emph{Proceedings of the IEEE/CVF International Conference on Computer Vision}, 2021, pp. 5752--5761.

\bibitem{DVGO}
C.~Sun, M.~Sun, and H.-T. Chen, ``Direct voxel grid optimization: Super-fast convergence for radiance fields reconstruction,'' in \emph{Proceedings of the IEEE/CVF Conference on Computer Vision and Pattern Recognition}, 2022, pp. 5459--5469.

\bibitem{vox-surf}
H.~Li, X.~Yang, H.~Zhai, Y.~Liu, H.~Bao, and G.~Zhang, ``Vox-surf: Voxel-based implicit surface representation,'' \emph{IEEE Transactions on Visualization and Computer Graphics}, 2022.

\bibitem{EG3D}
E.~R. Chan, C.~Z. Lin, M.~A. Chan, K.~Nagano, B.~Pan, S.~De~Mello, O.~Gallo, L.~J. Guibas, J.~Tremblay, S.~Khamis \emph{et~al.}, ``Efficient geometry-aware 3d generative adversarial networks,'' in \emph{Proceedings of the IEEE/CVF conference on computer vision and pattern recognition}, 2022, pp. 16\,123--16\,133.

\bibitem{iNGP}
T.~M{\"u}ller, A.~Evans, C.~Schied, and A.~Keller, ``Instant neural graphics primitives with a multiresolution hash encoding,'' \emph{ACM transactions on graphics (TOG)}, vol.~41, no.~4, pp. 1--15, 2022.

\bibitem{point-slam}
E.~Sandstr{\"o}m, Y.~Li, L.~Van~Gool, and M.~R. Oswald, ``Point-slam: Dense neural point cloud-based slam,'' in \emph{Proceedings of the IEEE/CVF International Conference on Computer Vision}, 2023, pp. 18\,433--18\,444.

\bibitem{GS-SLAM}
C.~Yan, D.~Qu, D.~Wang, D.~Xu, Z.~Wang, B.~Zhao, and X.~Li, ``Gs-slam: Dense visual slam with 3d gaussian splatting,'' \emph{arXiv preprint arXiv:2311.11700}, 2023.

\bibitem{3dgs}
B.~Kerbl, G.~Kopanas, T.~Leimk{\"u}hler, and G.~Drettakis, ``3d gaussian splatting for real-time radiance field rendering,'' \emph{ACM Transactions on Graphics}, vol.~42, no.~4, 2023.

\bibitem{replica}
J.~Straub, T.~Whelan, L.~Ma, Y.~Chen, E.~Wijmans, S.~Green, J.~J. Engel, R.~Mur-Artal, C.~Ren, S.~Verma \emph{et~al.}, ``The replica dataset: A digital replica of indoor spaces,'' \emph{arXiv preprint arXiv:1906.05797}, 2019.

\bibitem{scannet}
A.~Dai, A.~X. Chang, M.~Savva, M.~Halber, T.~Funkhouser, and M.~Nie{\ss}ner, ``Scannet: Richly-annotated 3d reconstructions of indoor scenes,'' in \emph{Proceedings of the IEEE conference on computer vision and pattern recognition}, 2017, pp. 5828--5839.

\bibitem{tum-rgbd}
J.~Sturm, N.~Engelhard, F.~Endres, W.~Burgard, and D.~Cremers, ``A benchmark for the evaluation of rgb-d slam systems,'' in \emph{2012 IEEE/RSJ international conference on intelligent robots and systems}.\hskip 1em plus 0.5em minus 0.4em\relax IEEE, 2012, pp. 573--580.

\bibitem{bundlefusion}
A.~Dai, M.~Nie{\ss}ner, M.~Zollh{\"o}fer, S.~Izadi, and C.~Theobalt, ``Bundlefusion: Real-time globally consistent 3d reconstruction using on-the-fly surface reintegration,'' \emph{ACM Transactions on Graphics (ToG)}, vol.~36, no.~4, p.~1, 2017.

\bibitem{meshlab}
P.~Cignoni, M.~Callieri, M.~Corsini, M.~Dellepiane, F.~Ganovelli, G.~Ranzuglia \emph{et~al.}, ``Meshlab: an open-source mesh processing tool.'' in \emph{Eurographics Italian chapter conference}, vol. 2008.\hskip 1em plus 0.5em minus 0.4em\relax Salerno, Italy, 2008, pp. 129--136.

\end{thebibliography}


\end{document}